\pdfoutput=1
\documentclass[twoside,twocolumn]{article}
\usepackage{blindtext} 

\usepackage[sc]{mathpazo} 
\usepackage[T1]{fontenc} 
\usepackage{microtype} 

\usepackage{cite}
\usepackage{amsmath,amssymb,amsfonts}
\usepackage{algorithmic}
\usepackage{graphicx}
\usepackage{extarrows}
\usepackage{multirow}
\usepackage{threeparttable}
\usepackage{array}
\usepackage{amssymb}
\usepackage{textcomp}
\usepackage{epstopdf}
\usepackage{threeparttable}
\usepackage{subfigure}

\usepackage[english]{babel} 

\usepackage[hmarginratio=1:1,left=18mm,right=18mm,top=32mm,columnsep=20pt]{geometry} 
\usepackage[hang, small,labelfont=bf,up,textfont=it,up]{caption} 
\usepackage{booktabs} 

\usepackage{lettrine} 

\usepackage{enumitem} 
\setlist[itemize]{noitemsep} 

\usepackage{abstract} 

\usepackage{titlesec} 
\renewcommand\thesection{\Roman{section}} 
\renewcommand\thesubsection{\roman{subsection}} 
\titleformat{\section}[block]{\large\scshape\centering}{\thesection.}{1em}{} 
\titleformat{\subsection}[block]{\large}{\thesubsection.}{1em}{} 

\usepackage{fancyhdr} 
\usepackage{titling} 

\usepackage{hyperref} 

\pretitle{\begin{center}\Huge\bfseries} 
\posttitle{\end{center}} 
\title{Asymmetric Residual Neural Network for Accurate Human Activity Recognition} 
\author{%
\textsc{Jun Long$^1$, WuQing Sun$^1$, Osolo Ian Raymond$^1$, Zhan Yang$^{1,2}$}\\[1ex] 
\normalsize $^1$School of Information Science and Engineering, Central South University, Changsha 410083, China \\ 
\normalsize $^2$Network Resources Management and Trust Evaluation Key Laboratory of Hunan Province \\ 
\normalsize \href{mailto:junlong@csu.edu.cn}{junlong@csu.edu.cn} 
\thanks{Accepted by Information, DOI:10.3390/info10060203}
}
\date{\today} 
\renewcommand{\maketitlehookd}{%
\begin{abstract}
Human Activity Recognition (HAR) using deep neural network has become a hot topic in human-computer interaction. Machine can effectively identify human naturalistic activities by learning from a large collection of sensor data. Activity recognition is not only an interesting research problem, but also has many real-world practical applications. Based on the success of residual networks in achieving a high level of aesthetic representation of the automatic learning, we propose a novel \textbf{A}symmetric \textbf{R}esidual \textbf{N}etwork, named ARN. ARN is implemented using two identical path frameworks consisting of (1) a short time window, which is used to capture spatial features, and (2) a long time window, which is used to capture fine temporal features. The long time window path can be made very lightweight by reducing its channel capacity, yet still being able to learn useful temporal representations for activity recognition. In this paper, we mainly focus on proposing a new model to improve the accuracy of HAR. In order to demonstrate the effectiveness of ARN model, we carried out extensive experiments on benchmark datasets (i.e., OPPORTUNITY, UniMiB-SHAR) and compared with some conventional and state-of-the-art learning-based methods. Then, we discuss the influence of networks parameters on performance to provide insights about its optimization. Results from our experiments show that ARN is effective in recognizing human activities via wearable datasets.
\end{abstract}
}

\begin{document}
\maketitle


\section{Introduction}
\label{sec:introduction}
Human Activity Recognition (HAR) is very important for human-computer interaction and is an indispensable part of many current real-world applications. To overcome the awareness of human-computer interaction, the potential features in the on-body device must be learned. HAR using wearable devices data is at the core of intelligent assistive technology, due to its proliferative applications in smart homes~\cite{b1}, intelligent traffic control~\cite{b2}, medical/health assistance~\cite{b3,b4}, skill based check~\cite{b5}, even in the security field~\cite{b6}. Particularly for the elderly people who are in remote and need to continuous monitor, HAR can greatly increase their safety~\cite{b7}.
\begin{figure*}
  \centering
  \includegraphics[scale=0.37]{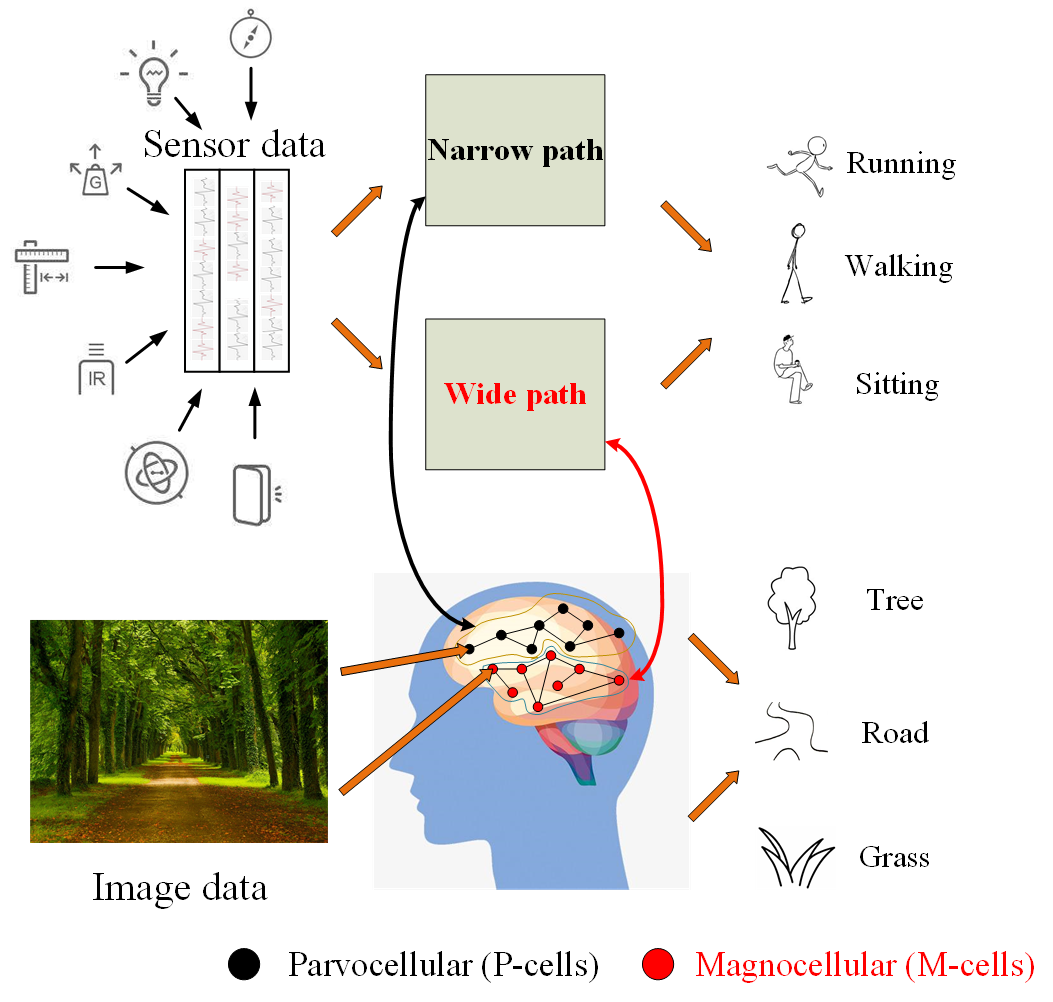}\\
  \caption{The basic motivation source of our proposed ARN.}\label{fig:introduction}
\end{figure*}

Nowadays, accelerators, gyroscopes and magnetic field sensors are widely utilized in smart phone (e.g., Apple iPhone, Samsung Galaxy, Huawei P/Mate), smart bracelets (Apple Watch, Fitbit). With the increasing number of wearable sensors and Internet of Things (IoT) devices, there is a growing trend in collecting the activity data of users in real time. The key technology in HAR includes a sliding time window of time-series data captured with on-body sensors, manually designed feature extraction procedures, and a wide variety of supervised learning methods.

In the past years, researchers have made lots of progress in wearable activity recognition, using algorithms such as Logistic Regression~\cite{b41}, Decision Tree~\cite{b9}, and Hidden Markov Model~\cite{b10}. Reference~\cite{b7} carried out experiment to evaluate the recognition performance of supervised and unsupervised machine learning techniques. For task identification, many of traditional methods (i.e., hand-crafted features and codebook approach) are characterized by manual feature extraction and discover features with expert knowledge. The performance of those methods often depends on the quality of the obtained feature representation. However, manual feature extraction is not always possible in practice, especially when we are unknown about the structure of the input data because of a lack of expert knowledge. Compared with manual feature extraction methods, deep learning techniques can discover adequate features without expert knowledge and systematic exploration of the feature space. In the many fields, deep learning techniques have achieved remarkable results, such as in image recognition~\cite{b11}, speech recognition~\cite{b12}, natural language processing~\cite{b13} and so on. Exiting deep learning methods for HAR can be further divided into two categories: Deep Neural Networks (DNNs)~\cite{b14}, and Convolutional Neural Networks (CNNs)~\cite{b15}. DNNs (e.g., contain an input layer, at least two hidden layers and an output layer) can extract highly abstract features by stacking hidden layers. It allows us to add more possible connections between input and output neurons so that the ways to re-use learned features can be increased. Researchers who have used DNNs methods for HAR include:~\cite{b16} who investigated deep neural networks with wearable sensors data and~\cite{b17} who explored temporal deep neural networks for active biometric authentication.

Recently, many researchers have adopted CNNS to deploy human activity recognition system, such as~\cite{b6,b16,b18,b19}. CNNs can model the entire sequence by sharing the weights from local to global, extract abstract features at hierarchical layers through a series of convolutional operations, and process the raw activity signals for capturing potential features. CNNs are based on the discovery of visual cortical cells and retain the spatial information of the data through receptive field. It is known that the power of CNNs stems in large part from their ability to exploit symmetries through a combination of weight sharing and translation equivariance. Also, with their ability to act as feature extractors, a plurality of convolution operators is stacked to create a hierarchy of progressive abstract features. Reference~\cite{b15} proposed CNNs based approaches to automatically extract discriminative features for HAR. More and more researches are using variants of CNNs to learn sensor-based data representations for human activity recognition and have achieved remarkable performances. The model~\cite{b20} consists of two or three temporal-convolution layers with a ReLU activation function followed by a max-pooling layer and a soft-max classifier, which can be applied over all sensors simultaneously. Yang \emph{et al.}~\cite{b18} introduce four temporal-convolutional layers on a single sensor, followed by a fully-connected layer and soft-max classifier and it shows that deeper networks can find correlations between different sensors.

However, all the deep learning methods we mentioned above are all identified by a single-path neural net without considering spatial and temporal features of the data. Inspired by the biological study of retinal ganglion cells in the primate visual system (Figure~\ref{fig:introduction} illustrates the basic motivation source of our proposed ARN), there are Parvocellular (P-cells) that provide good spatial detail and color in the visual system, but its resolution is very low. In addition, there are high-frequency Magnocellular (M-cells), which are very sensitive to time changes, but not sensitive to spatial details and colors. In this paper, inspired by the facts above, we propose a model to handle a set of activity data, synchronized by an asymmetric net, using a short time window to capture spatial features and a long time window to capture fine temporal features, corresponding to the P-cells and the M-cells, respectively. Our network is an end-to-end network, and the input of the network is the original sensor data. The data collected from the wearable device can be directly input into the network. ARN model is applicable for supervised learning approaches and unsupervised learning approaches because it is based on ResNet~\cite{b21} that has been proven to be applicable for supervised and unsupervised learning approaches. In this paper, we use supervised learning method. Because it can make label information bridge the heterogeneous gap. 

We propose a novel Asymmetric Residual Network, named ARN. As a new kind of deep learning network, the components of activity recognition in ARN are divided into two parts. (1) a residual net using short time window (i.e., 32 or 64); (2) a residual net using long time window (i.e., 64 or 96). 32, 64, 96 are meaning the length of the time window\footnote{(Length) = (Time) x (Sampling Frequency)}. The last layer representations of two parts will be concatenated, then use the fusion representations for accurate activity recognition. The superior advantages of the ARN over other existing methods were listed in the Table~\ref{tab:Comparison with other methods}. To the best of our knowledge, this is the first work that applies a asymmetric residual net for activity recognition.

The main contributions of the paper are as follows:
\begin{enumerate}
  \item We propose a novel symmetric neural network based on ResNet for HAR, termed ARN, which is an asymmetric network and has two paths separately working at short and long slide window, our wide path is designed to capture global features but few spatial details, analogous to M-cells, and our narrow path is lightweight, similar to the small receptive field of P-cells.
  \item We design a network that consists of asymmetric residual net that not only can effectively manage information flow, but will also automatically learn effective activity feature representation, while capturing the fine feature distribution in different activities from wearable sensor data. 
  \item We compare the performance of our method with other relevant methods by carrying out extensive experiments on benchmark datasets. The results show our method outperforms other methods.
\end{enumerate}

The remainder of this paper is structured as follows. In Section~\ref{sec:rw}, we briefly introduce the related works. In Section~\ref{sec:method}, we highlight the motivation of our method and provide some theoretical analysis for its implementation. In Section~\ref{sec:exp}, we introduce our experimental results and corresponding analysis and finally in Section~\ref{sec:conc} concludes the paper.

\begin{table*}[tp]
  \centering
  \begin{threeparttable}
    \centering
    \caption{Comparison of different HAR technologies}
    \label{tab:Comparison with other methods}
      \begin{tabular}{c|c|c|c|c|c|c}
      \toprule
      \specialrule{0em}{2pt}{2pt}
      Method&manual f.$^1$&high-level f.$^1$&spatial f.$^1$&temporal f.$^1$&unsupervised&supervised\cr
      \midrule
      \specialrule{0em}{2pt}{2pt}
      HC~\cite{b22}&\checkmark&$\times$&$\times$&$\times$&\checkmark&$\times$\cr
      \specialrule{0em}{2pt}{2pt}
      CBH~\cite{b23}&\checkmark&$\times$&$\times$&$\times$&\checkmark&$\times$\cr
      \specialrule{0em}{2pt}{2pt}
      CBS~\cite{b24}&\checkmark&$\times$&$\times$&$\times$&\checkmark&$\times$\cr
      \specialrule{0em}{2pt}{2pt}
      AE~\cite{b25}&\checkmark&$\times$&$\times$&$\times$&\checkmark&$\times$\cr
      \specialrule{0em}{2pt}{2pt}
      MLP~\cite{b26}&\checkmark&$\times$&$\times$&$\times$&\checkmark&$\times$\cr
      \specialrule{0em}{2pt}{2pt}     
      CNN~\cite{b15}&$\times$&\checkmark&$\times$&$\times$&\checkmark&\checkmark\cr
      \specialrule{0em}{2pt}{2pt}
      LSTM~\cite{b27}&$\times$&\checkmark&$\times$&$\times$&$\times$&\checkmark\cr
      \specialrule{0em}{2pt}{2pt}
      Hybrid~\cite{b28}&\checkmark&\checkmark&$\times$&$\times$&$\times$&\checkmark\cr
      \specialrule{0em}{2pt}{2pt}
      ResNet~\cite{b21}&\checkmark&\checkmark&$\times$&$\times$&\checkmark&\checkmark\cr
      \specialrule{0em}{2pt}{2pt}
      \hline
      \specialrule{0em}{2pt}{2pt}
      \textbf{ARN(This Work)}&\checkmark&\checkmark&\checkmark&\checkmark&\checkmark&\checkmark\cr
      \specialrule{0em}{2pt}{2pt}
      \bottomrule
      \end{tabular}
        \begin{tablenotes}
        \item[1] f. denotes features.
        \end{tablenotes}
  \end{threeparttable}
  \end{table*}

\section{Related Work}\label{sec:rw}

\subsection{Methods for Human Activity Recognition}
This section introduces the features extraction methods for HAR selected in our comparative study. There are two main directions for HAR methods: conventional recognition methods and learning-based methods. In conventional methods, we extract features manually with expert knowledge. In Learning-based methods, we can discover adequate features without expert knowledge and systematic exploration of the feature space. Conventional methods in our comparative study include Hand-Crafted Features (HC)~\cite{b22}, Codebook approach (CB)~\cite{b23}. The learning-based methods include Autoencoders approach (AE)~\cite{b25}, Multi-Layer Perceptron (MLP)~\cite{b26}, Convolutional Neural Network (CNN)~\cite{b15}, Long-Short Term Memory Networks (LSTM)~\cite{b27}, Hybrid Convolutional and Recurrent Networks (Hybrid)~\cite{b28}, Deep Residual Learning (ResNet)~\cite{b21}.
\subsubsection{Hand-Crafted Features}
HC comprises simple metrics computed on data and uses simple statistical value (e.g., std, avg, mean, max, min, median, etc.) or frequency domain correlation features based on the signal Fourier transform to analyze the time series of human activity recognition data. Due to its simplicity to setup and low computational cost, it is still being used in some areas, but the accuracy cannot satisfy the requirement of modern AI games. In addition, when faced with the activity recognition of complex high-level behaviors tasks, identifying the relevant features through these traditional approaches is time-consuming.
\subsubsection{Codebook approach}
CB consists of two consecutive steps. The first step is codebook construction which is to construct codebook by using cluster algorithm to process a set of subsequences extracted from the original data sequence. Each center of the cluster is considered as a codeword which represents a distinct subsequence. The second step is codeword assignment which aims to built a feature vector that is associated to a data sequence. Subsequences are firstly extracted from the sequence, and then assign each subsequence to the most similar codeword. Finally, a histogram-based feature representing the distribution of all codewords is built by using this information.
During the codebook construction, a set of subsequences can be firstly extracted from the original data sequence by using a sliding time window approach with window size $w$ and sliding stride $h$. Then, a k-means clustering~\cite{b29} algorithm can be applied on subsequences to obtain $n$ clusters of similar subsequences. Similarity metric between two subsequences is Euclidean distance. Finally, we can get a codebook that consists of $n$ codewords.

During the codebook assignment, we should firstly extract subsequences from a sequence using the same sliding window approach as the codebook construction, and the most similar codeword need to be assigned for each subsequence.
Then, a histogram of the frequencies of codewords can be built by using this information. Finally, we can get a probabilistic feature presentation by normalizing the histogram.

The approach of described above can be called codebook with hard assignment (CBH) because each subsequence is assigned to a codeword deterministically. However, this approach may lack flexibility in some uncertain situations where  a subsequence is similar to two or more codewords. In order to solve this problem, we use a soft assignment variant (CBS). CBS can exploit kernel density estimation~\cite{b24} to perform smooth assignment of subsequences to multiple codewords. It allows us to obtain a feature which represents a smooth distribution of codewords that considers the similarity between all codewords and subsequences.
\begin{figure*}
  \centering
  \includegraphics[scale=0.43]{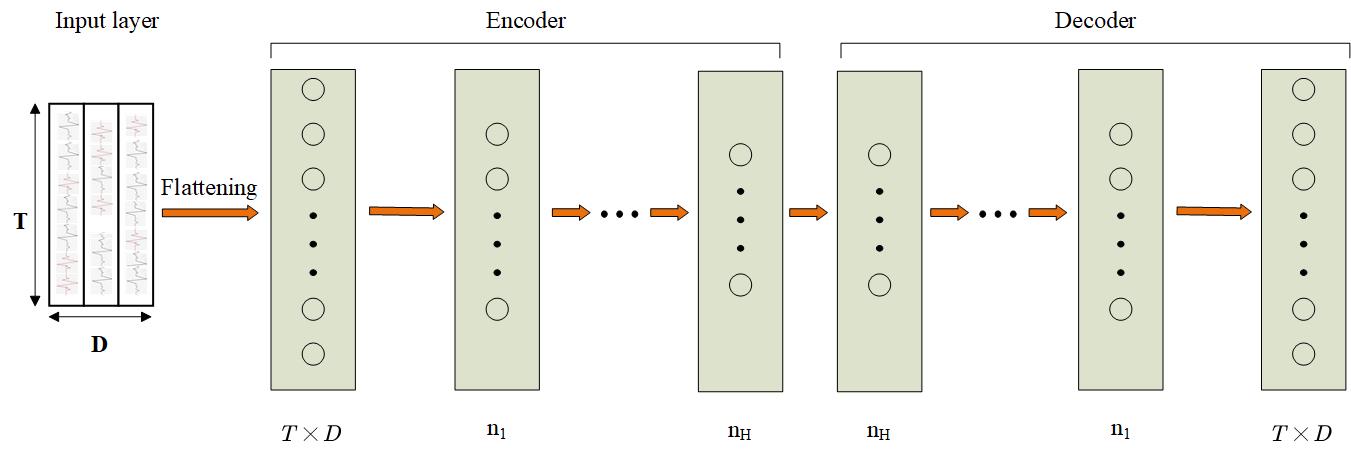}\\
  \caption{Architecture of a AE for HAR. $T$ in the input layer denotes the length of time window. $D$ denotes the number of sensor channels. $H$ denotes the number of hidden layers.}\label{fig:autoencoder}
\end{figure*}   
\subsubsection{Autoencoders approach}
AE is a specific architecture that consists an encoder and a decoder as depicted in Fig~\ref{fig:autoencoder}. The encoder can project input data in a feature space of lower dimension. While the decoder can map the encoded features back to the input space. Then, the AE can reproduce the input data on the output according to a loss function like Mean Squared Errors after t
raining. 
\subsubsection{Multi-Layer Perceptron}
\begin{figure*}
  \centering
  \includegraphics[scale=0.43]{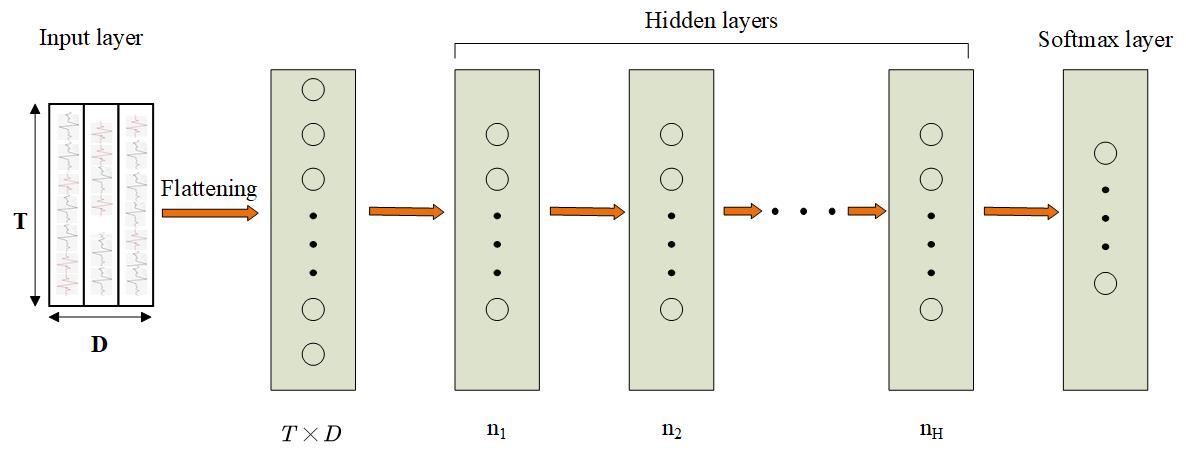}\\
  \caption{Architecture of a MLP for HAR. $T$ in the input layer denotes the length of time window. $D$ denotes the number of sensor channels. $H$ denotes the number of hidden layers. Input data are first flattened into a ($T\times D$)-dimensional vector. The vector is used as data for hidden layers. All layers are fully-connected.}\label{fig:mlp}
\end{figure*}

MLP is one of the most simplest neural networks. An important feature of the MLP is that it has multiple layers. As show in Fig~\ref{fig:mlp}, we call the first layer as input layer, the last layer as output layer, and the middle layers as hidden layers. MLP does not specify the number of hidden layers, so you can choose the appropriate number of hidden layers according to your needs. Each neuron in a fully-connected layer takes the outputs of the previous layer as its inputs. Stacking layers can be seen as extracting features of an increasingly higher level of abstraction and the output features of neurons at $n$-th layer can be calculated by neurons at $(n-1)$-th layer.
\subsubsection{Convolutional Neural Networks}
CNNs can automatically extract the features from raw sensor data which without need for very professional expert knowledge~\cite{b30}. A standard convolutional neural network consists of convolutional layers, max-pooling layers, fully-connection layers (FC) and a Soft-Max layer. Instead of using predefined filters as in traditional feature extracting methods, CNNs can learn locally connected neurons that represent data-specific filters. As CNNs can share weights of neurons, the parameters of CNNs are much fewer than those of the traditional neural networks~\cite{b31}.\\

\noindent\textbf{Convolutional layers} are an important component of CNNs. Using several convolution filters (or kernels), which aim to learn feature representations of the raw input, complex operations can be easily performed by the convolution operation in the convolutional layer. The dimension of filters (or kernels) is determined by the input dimension. Convolution kernel is a function that generalizes a linear model for the underlying local patch. It works well for abstraction, when instances of latent concepts are linearly separable. In each convolutional layer, neurons of current layer are connected to the neurons of previous layer through feature mapping operation. Thus, feature mapping of the upper layer can be obtained from the convolved results of the previous layer by adopting an element-wise nonlinear activation function. So, the value of the feature map $j$ in the $l$-th layer, $x_j^{l+1}$  is calculated by:
\begin{equation}\label{eq:rw_1}
  x_j^{l+1}=\sigma(\sum_{i\in\text{Maps}}x_i^l\otimes K_{ij}^l+b_j^l)
\end{equation}
where $\text{maps}$ are the total number of feature maps in $l$-th layer and $b_j^l$ is a bias vector. $\sigma(\cdot)$ is the activation function to improve the performance of CNNs. The most notable non-liner activation function is \emph{ReLU}, which is defined as: $\sigma(x)=\text{max}(x,0)$. The \emph{ReLu} activation operation allows networks to compute much faster than \emph{sigmoid} or \emph{tanh} activation functions, induces the sparsity in the hidden neurons, and makes networks to obtain sparse representations more easily. Adopting ReLU may bring zero value to affect the performance of backpropagation, but many research results have show that \emph{ReLU}~\cite{b32} works much better than \emph{sigmoid} and \emph{tanh}~\cite{b33}.\\ 

\noindent\textbf{Pooling layers} have come after the convolutional layer, is another component of CNNs. In the pooling layer, a pooling operation is used to reduce the number of neurons connections between neighboring convolutional layers thus reducing computational complexity.\\

\noindent\textbf{Fully-connected layers}, whcih aims to convert the matrix-feature (2-D) unfolded to a vector-feature (1-D) for anastomosis classification tasks, and contains about 90\% of the parameters of the entire CNNs.\\

\noindent\textbf{Loss function} plays an important role in different classification tasks. The most common loss function is soft-max. Given a training set $\{x^{(i)},y^{(i)};i\in[1,N],y^{(i)}\in[1,K]\}$ , where $x^{(i)}\in\mathbb{R}^D$ is the input patch, $y^{(i)}$ is the target label which belongs to the total number of labels (K). The prediction $a_j^{(i)}$ of $j$-th class for $i$-th input is transformed with the Soft-max function:
      \begin{equation}\label{eq:soft-max}
        p(a^{(i)}=j)=\frac{e^{a_j^{(i)}}}{\sum_{K}e^{a_k^{((i))}}}
      \end{equation}
      Soft-max normalizes the predictions to a probability distribution over the total classes. The soft-max is represented loss as follows:
      \begin{equation}\label{eq:soft-max-loss}
        \ell=-\frac{1}{N}(\sum_{l=1}^{N}\sum_{j=1}^{K}1\{y^{(i)}=j\}\cdot\log p(a^{(i)}=j))
      \end{equation}\\

\noindent\textbf{Regularization} is required in CNNs.  Overfitting is an unavoidable problem in convolutional neural networks, that but it can be effectively reduced by regularization. As a means of regularization, dropout can prevent the dependence of different neurons in a network, and force the network to be more accurate even in the absence of certain information.\\

\begin{figure*}
  \centering
  \includegraphics[scale=0.39]{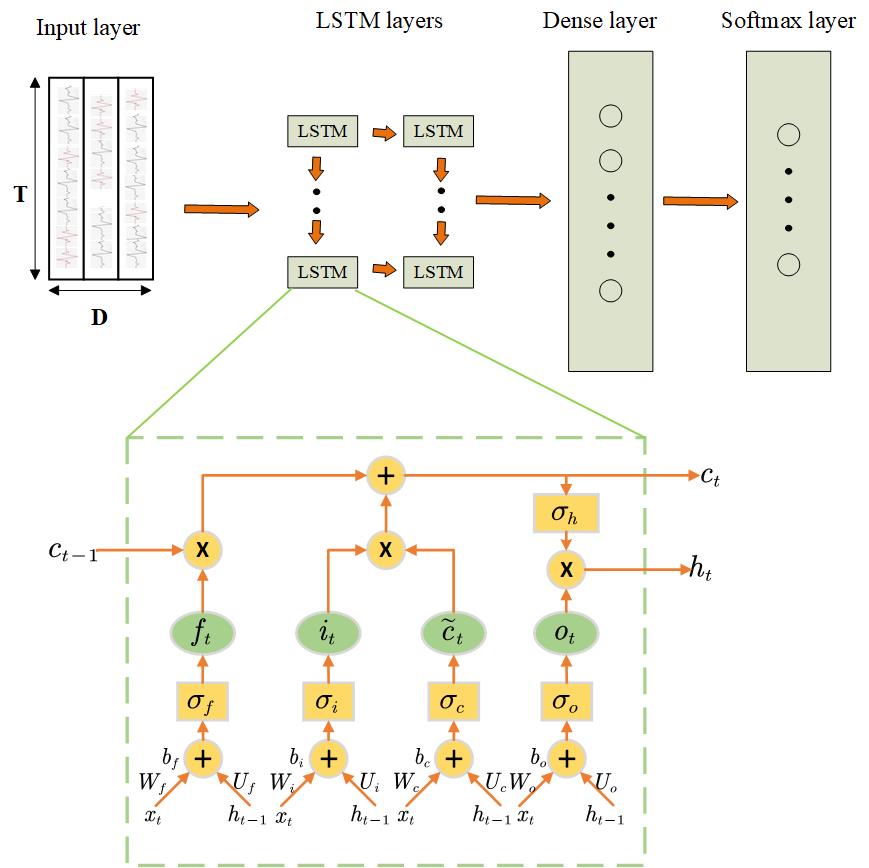}\\
  \caption{Architecture of a LSTM for HAR. $T$ in the input layer denotes the length of time window. $D$ denotes the number of sensor channels. Each LSTM layer is composed of LSTM cell. $\otimes$ denotes the element-wise multiplication. $\oplus$ denotes the element-wise addition. The output of the last LSTM layer is passed to a dense and a Soft-Max layers.}\label{fig:lstm}
\end{figure*}
\subsubsection{Recurrent Neural Networks and Long-Short Term Memory Networks}
Recurrent Neural Networks (RNNs) are a specific architecture that connections between neurons have directed cycles and the output of the neurons dependent on the state of the network at the previous timestamps. RNNs can find patterns with long-term dependencies because its specific behavior that memorizes the information extracted from the past data. But in practice, there is a phenomenon called gradient explosion or vanish will make a great affect on performance of RNNs. The problem of vanishing or exploding gradient refers to the derivate of the error function with respect to the network weight becomes very large or close to zero~\cite{b34}. This problem will result in the adverse impact on the weight update by the back-propagation algorithm. Therefore, LSTM is designed to solve the problem of vanishing or exploding gradient in RNNs. LSTM extends RNNs with memory cells and remembers information over time by storing it in an internal memory. The internal state can be updated and erased depending on their input and the state at the previous time step. As show in Fig~\ref{fig:lstm}, this mechanism is achieved by introducing internal processor called cell. A cell contains three gates, called input gate ($i_t$), output gate ($o_t$) and forget gate ($f_t$). $c_t$ is the cell state. Gates are used to regulate the information update to the cell state. Their equations are mentioned below:
\begin{equation}\label{eq:rw_1}
  f_t=\sigma _f \left( W_fx_t+U_fh_{t-1}+b_f \right)
\end{equation}
\begin{equation}\label{eq:rw_1}
  i_t=\sigma _i \left( W_ix_t+U_ih_{t-1}+b_i \right) 
\end{equation}
\begin{equation}\label{eq:rw_1}
  o_t=\sigma _o\left( W_ox_t+U_oh_{t-1}+b_o \right) 
\end{equation}
\begin{equation}\label{eq:rw_1}
  \tilde{c}_t=\sigma _c\left( W_cx_t+U_ch_{t-1}+b_c \right) 
\end{equation}
\begin{equation}\label{eq:rw_1}
  c_t=f_t\cdot c_{t-1}+i_t\cdot \tilde{c}_t
\end{equation}
\begin{equation}\label{eq:rw_1}
  h_t=o_t\cdot \sigma _h\left( c_t \right) 
\end{equation}
where $\cdot$ represents the element-wise multiplication of two vectors. $x_t$ refers to the input vector to the LSTM cell at time t and $h_t$ is the hidden state vector. $\sigma$ designates the activation function. $W$, $U$ and $b$ are the matrices of weights and biases.
\subsubsection{Hybrid Convolutional and Recurrent Networks}
\begin{figure*}
  \centering
  \includegraphics[scale=0.43]{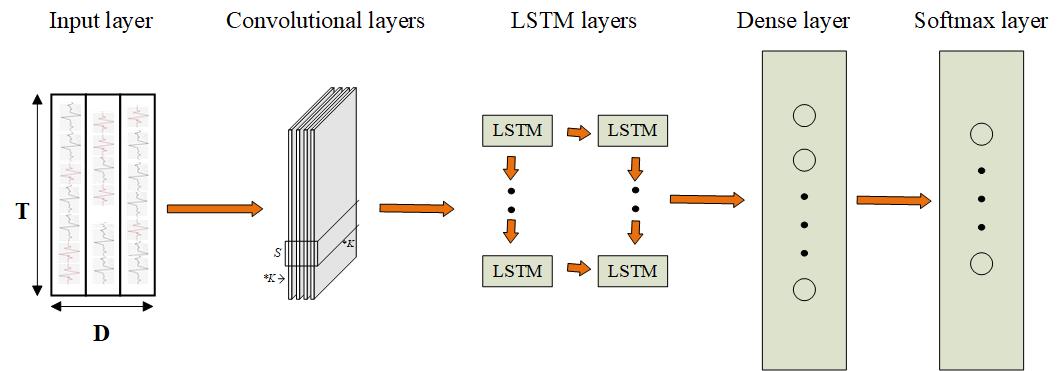}\\
  \caption{Architecture of a hybrid for human activity recognition. $T$ in the input layer denotes the length of time window. $D$ denotes the number of sensor channels. In convolutional layers, $K$ denotes the kernels in layer and $S$ is the length of kernels. The convolutional layers can extract features from input data and provide abstract presentations of data in feature maps. LSTM layers learn temporal dynamics from data.}\label{fig:hibrid}
\end{figure*}

Hybrid comprises convolutional, LSTM and softmax layers as depicted in Fig~\ref{fig:hibrid}. Convolutional layers have the ability to extract the features from input data and create a hierarchy of progressively more abstract features by stacking several convolutional operators. LSTM includes a memory to model temporal dependencies in time series problems. Therefore, the combination of CNN and LSTM can capture time dependencies on features extracted by convolutional operations.    
\subsubsection{Deep Residual Learning}
ResNet is designed to address degradation problem that the accuracy of training set gets saturated or even decreases with the network depth increasing. Different from the ordinary convolutional neural network, ResNet has many stacked Residual blocks, in which identity mappings are added to connect input and output. Residual block with identity mapping can be expressed in a general form:
\begin{equation}\label{eq:rw_1}
  y_l=h\left( x_l \right) +F\left( x_l,W_l \right) 
\end{equation}
\begin{equation}\label{eq:rw_1}
  x_{l+1}=f\left( y_l \right)  
\end{equation}
where $x_{l+1}$ and $x_l$ are output and input of the $l$-th unit, $F$ is a residual function and $W_l$ are parameters of the unit. $h\left( x_l \right) =x_l$ is an identity mapping and $f$ is a ReLU activation function. The key idea of ResNet is to learn the additive residual function $F$ with respect to $h\left( x_l \right)$. ResNet can make the element-wise addition on input and output by attaching a shortcut connection. This simple addition can increase the training speed of the model and improve the training effect, and will not add additional parameters to the network. With the network depth increasing, this simple structure is a good solution to the degradation problem. 

\section{Asymmetric Residual Network}\label{sec:method}
In this section, our proposed model has a narrow path (see Sec~\ref{sec:sp}) and a wide path(see Sec~\ref{sec:hp}), which are concatenated and sent to the fully-connected layer. Loss function is introduced in Sec~\ref{sec:loss}.

\begin{figure*}
    \centering
    \includegraphics[scale=0.43]{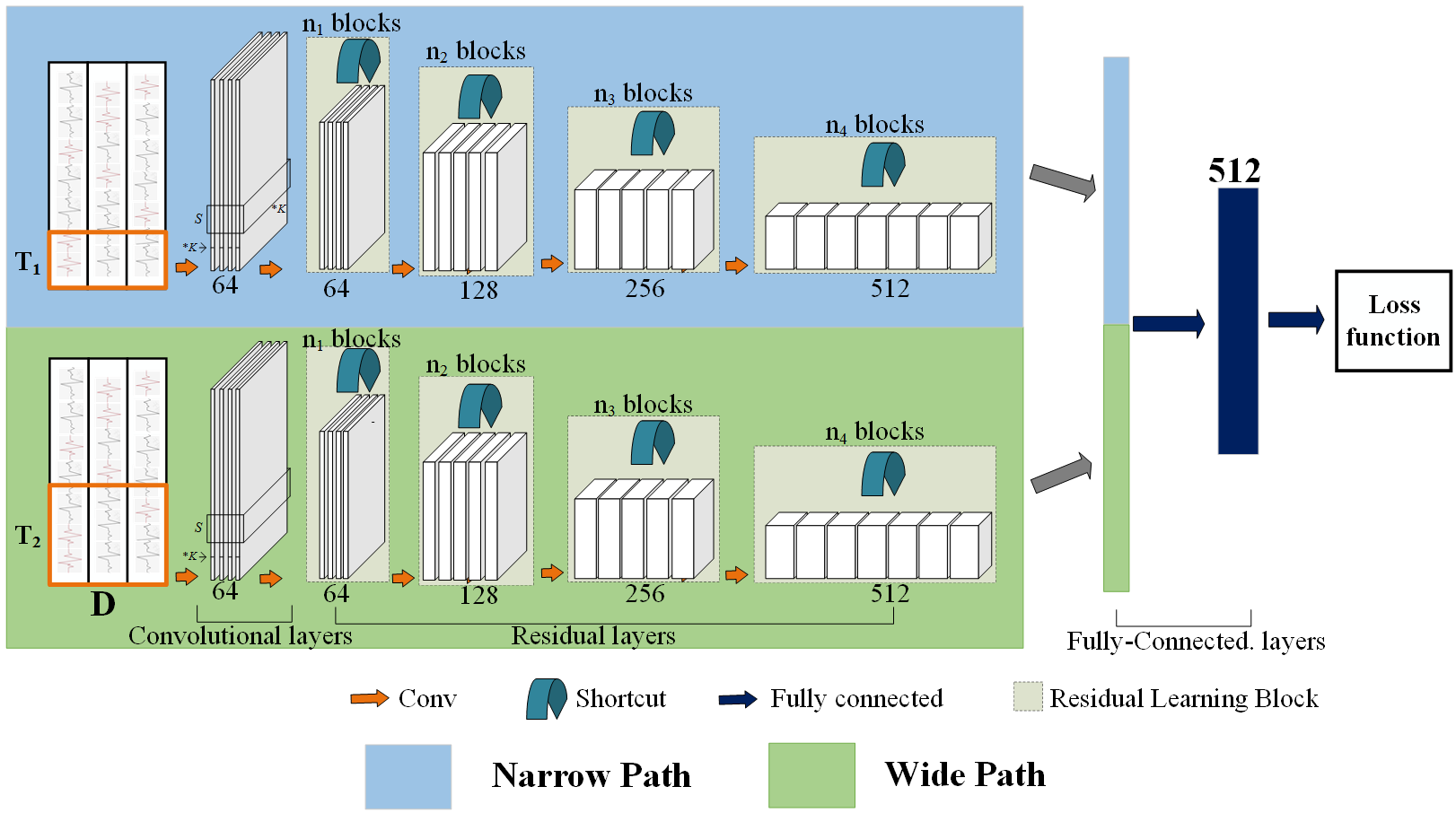}\\
    \caption{The proposed ARN architecture for human activity recognition. $T_1$ and $T_2$ in the raw data layer denote the length of time window corresponding to the narrow path and wide path, respectively ($T_1<T_2$). $D$ denotes the number of sensor channels. In the conv. layers, $K$ denotes the kernels in layer, and the length of kernels is $s=5$. In the res. layers, the dimension of each block double increase feature maps to the input signal, which is processed by number of preset building blocks for residual learning.}\label{fig:overall}
  \end{figure*}

\subsection{Network Architecture}\label{sec:na}
As shown in Fig~\ref{fig:overall}, convolutional layers and residual layers in our architecture are used to model the recognition task. In convolutional layers, the general activity features are extracted from raw sensor data. In residual layers, the special features can be extracted from general features and the special features are used for human activity recognition.

The convolutional layers (i.e., conv. layers) of our architecture consists of 1 layer, including 64 sliding windows (filters) whose size is $s=5\times 1$ , a batch normalization layer, and a ReLU layer with use of a pooling layer~\cite{b30}. The residual layers contain four “blocks”. The details of residual block are shown in Table~\ref{tab:model_para} and the value of $n_1,n_2,n_3,n_4$  are set to $3,4,6,3$, respectively.

\subsection{Narrow Path}\label{sec:sp}
The narrow path can be any convolutional model (e.g., Reference~\cite{b35} introduced a new Two-Stream inflated 3D Convolutional Networks: filters and pooling kernels of very deep image classification Convolutional Networks are expanded into 3D, Reference~\cite{b36} introduced spatiotemporal ResNets as a combination of Two-stream Convolutional Networks and ResNets, Reference~\cite{b37} introduced non-local operations as a generic family of building blocks for capturing long-range dependencies.) that works on a sequence data as a spatiotemporal volume. The key concept in our narrow path is a short slide time window to scan the sequence activity data. We can know form~\cite{b38} that feature learning methods can get good performance when $T=32, 64, 96$. Therefore, a typical value of $T$ we study is 32~\cite{b38}. Denoting the number of sensor channels as $S$, the raw clip length is $T\times S$. The function of this path is to throw compact information into the net, the purpose is to capture spatial features.

\subsection{Wide Path}\label{sec:hp}
In parallel to the narrow path, the wide path is another convolutional model with a long slide time window. The operations of two path net work on the same raw activity data sequences, so the wide path uses $\alpha T$ slide time window, $\alpha$ times longer than the narrow path. A typical value is $\alpha=3$~\cite{b39} in our experiments. The presence of $\alpha$ is in the key of the NarrowWide concept. It explicitly indicates that the two paths work on different time window. Our wide path enters a long sequence of activity data into the net in order to pursue global functionality throughout the net hierarchy. Our wide path is distinguished from existing methods in that it can use significantly lower channel capacity to achieve good accuracy for the ARN model. The low channel capacity can also be interpreted as a weaker ability of representing spatial semantics. Our wide path not only has a long slide time window, but also pursues high-dimension features throughout the network hierarchy, maintaining temporal fidelity as much as possible.

\subsection{Lateral Concatenation}
Our lateral concatenation fuses from the narrow path to the wide path. We denote the representation shape of the narrow path as $\{T,S\}$, the representation shape of the wide path is $\{\alpha T, S\}$. The output of the lateral concatenation is fused into the narrow path by concatenation. Therefore, the shape of the concatenation layer is $\{(1+\alpha)T,S\}$.

\subsection{Loss Function}\label{sec:loss}
In order to train classification models, classification objectives (such as logistic loss and softmax loss) have been widely explored. For accurate human activity recognition, using labels that are different from the ground-truth for prediction, cannot contribute to the update of the network parameters. For depth estimates, predictions that are close to the ground-truth labels also help to update network parameters. In this work, we employ softmax loss for training the human activity recognition model. For each training sequence $x$, the probability of each label $k\in \{1,2,...,\mathcal{K}\}$ in our model is computed via softmax:
\begin{equation}\label{eq:softmax}
  p(k|x)=\frac{exp\{z_k\}}{\sum_{i=1}^{\mathcal{K}}exp\{z_i\}}
\end{equation}
where $z_i$ are the logits or unnormalized log probabilities. Here, the $z_i$ are computed by adding a fully-connected layer on top of the sequence data embedding, i.e., $z_i=W_i^T\phi(x)+b_i$, where $W_i$ and $b_i$ are weights and bias for target label, respectively. Let $q(k|x)$ denote the ground-true distribution over classes for this training example such that $\sum_{i=1}^{\mathcal{K}}q(k|x)=1$. The cross-entropy loss for the example is computed as:
\begin{equation}\label{eq:loss}
  \ell=-\sum_{k=1}^{\mathcal{K}}log(p(k|x))q(k|x)
\end{equation}


\begin{table*}[tp]
\centering
\begin{threeparttable}
  \centering
  \caption{The layer-parameters of the ARN mdoel.}
  \label{tab:model_para}
    \begin{tabular}{ccccc}
    \toprule
    Stage&narrow path&wide path\cr
    \midrule
    \specialrule{0em}{2pt}{2pt}
    Conv1&$64\times(5,1)$&$64\times(5,1)$\cr
    \specialrule{0em}{2pt}{2pt}
    \hline
    \specialrule{0em}{2pt}{2pt}
    Max-pooling&/2&/2\cr
    \specialrule{0em}{2pt}{2pt}
    \hline
    \specialrule{0em}{2pt}{2pt}
    res1&$\begin{bmatrix}
            1\times1 & 64 \\
            3\times1 & 64 \\
            1\times1 & 256
          \end{bmatrix}\times 3$&$\begin{bmatrix}
            1\times1 & 64 \\
            3\times1 & 64 \\
            1\times1 & 256
          \end{bmatrix}\times 3$\cr
    \specialrule{0em}{2pt}{2pt}
    \hline
    \specialrule{0em}{2pt}{2pt}
    res2&$\begin{bmatrix}
            1\times1 & 512 \\
            3\times1 & 512 \\
            1\times1 & 2048
          \end{bmatrix}\times 4$&$\begin{bmatrix}
            1\times1 & 512 \\
            3\times1 & 512 \\
            1\times1 & 2048
          \end{bmatrix}\times 4$\cr
    \specialrule{0em}{2pt}{2pt}
    \hline
    \specialrule{0em}{2pt}{2pt}
    res3&$\begin{bmatrix}
            1\times1 & 256 \\
            3\times1 & 256 \\
            1\times1 & 1024
          \end{bmatrix}\times 6$&$\begin{bmatrix}
            1\times1 & 256 \\
            3\times1 & 256 \\
            1\times1 & 1024
          \end{bmatrix}\times 6$\cr
    \specialrule{0em}{2pt}{2pt}
    \hline
    \specialrule{0em}{2pt}{2pt}
    res4&$\begin{bmatrix}
            1\times1 & 128 \\
            3\times1 & 128 \\
            1\times1 & 512
          \end{bmatrix}\times 3$&$\begin{bmatrix}
            1\times1 & 128 \\
            3\times1 & 128 \\
            1\times1 & 512
          \end{bmatrix}\times 3$\cr
    \specialrule{0em}{2pt}{2pt}
    \hline
    \specialrule{0em}{2pt}{2pt}
    concate&global average pool&global average pool\cr
    \specialrule{0em}{2pt}{2pt}
    \hline
    \specialrule{0em}{2pt}{2pt}
    fc.&512&512\cr
    \bottomrule
    \end{tabular}
\end{threeparttable}
\end{table*}

\section{Experiments}\label{sec:exp}
In order to demonstrate the performance of our proposed ARN method, we carried out our extensive experiments on two widely used benchmark datasets, i.e., OPPORTUNITY and UniMiB-SHAR, to verify the effectiveness of our method.

\subsection{Dataset}
Human activity features are usually unique and cyclical, and natural human activities include walking, running, jumping and so on. Therefore, a set of active data that includes a variety of types of natural human activities should be considered in dataset construction.

We use benchmark datasets to validate the model performance, and use different action sequences to verify whether they belong to the same person. There are many benchmark activity datasets, such as OPPORTUNITY~\cite{b40}, WISDM~\cite{b41}, UniMiB-SHAR~\cite{b42}, MHEALTH~\cite{b43}, PAMAP2~\cite{b44} datasets. In this paper, we evaluate our method by using the following two datasets.

\textbf{OPPORTUNITY} dataset has been widely used in many researches. It contains four subjects performing 17 different (morning) Activities of Daily Living (ADLs) in a sensor-rich environment, as listed in Table~\ref{tab:details}~\ref{tab:OPPORTUNITY_DATASET}. They were acquired at a sampling frequency of 30Hz equipping 7 wireless body-worn inertial measurement units (IMUs). Each IMU consists of a 3D accelerometer, 3D gyroscope and a 3D magnetic sensor, as well as 12 additional 3D accelerometers placed on the back, arms, ankles and hips, accounting for a total of 145 different sensor channels. During the data collection process, each subject performed a session 5 times with ADL and 1 drill session. During each ADL session, subjects were asked to perform the activities naturally-named ``ADL1'' to ``ADL5''. During the drill sessions, subjects performed 20 repetitions of each of the 17 ADLs of the dataset. The dataset contains about 6 hours of information in total, and the data are labeled on a timestamp level. The dataset can be used in an open activity recognition recognition challenge where participants competed to achieve the highest performance on the recognition. In our experiment, the training and testing sets have \textbf{63}-\textbf{D}imensions (\textbf{36}-\textbf{D} on hand, \textbf{9}-\textbf{D} on back and \textbf{18}-\textbf{D} on ankle, respectively).

\textbf{UniMiB-SHAR} dataset was collected data from 30 healthy subjects (6 male and 24 female) acquired using the 3D-accelerometer of a Samsung Galaxy Nexus I9250 with Android OS version 5.1.1. It contains 11771 samples of both human activities and falls performed by 30 subjects of ages ranging from 18 to 60. The data are sampled at a constant sampling rate of 50 Hz, and split into 17 different activity classes, 9 safety activities and 8 dangerous activities (e.g., a falling action) as shown in Table~\ref{tab:details}~\ref{tab:UniMiB-SHAR_DATASET}. Unlike the OPPORTUNITY dataset, the dataset does not have any NULL class and remains relatively balanced. It allows researchers to work to more robust features and classification schemes. In our experiments, the training and testing sets have \textbf{3}-\textbf{D}imensions.

\newcommand{\tabincell}[2]{\begin{tabular}{@{}#1@{}}#2\end{tabular}}
\begin{table*}[tp]
  \centering
  \begin{threeparttable}
    \centering
    \caption{The details of experimental datasets}
    \label{tab:details}
      \begin{tabular}{c|c|c}
      \toprule
      \specialrule{0em}{2pt}{2pt}
      Datasets&OPPORTUNITY&UniMiB-SHAR\cr
      \midrule
      \specialrule{0em}{2pt}{2pt}
      Types of sensors& \tabincell{c}{Custom bluetooth wireless accelerometers, \\gyroscopes, \\Sun SPOTs and InertiaCube3, \\Ubisense localisation system, \\ A custom-made magnetic field sensor}&A Bosh BMA220 acceleration sensor\cr
      \specialrule{0em}{2pt}{2pt}
      \hline
      \specialrule{0em}{2pt}{2pt}
      Numbers of sensors&72&3\cr
      \specialrule{0em}{2pt}{2pt}
      \hline
      \specialrule{0em}{2pt}{2pt}
      Numbers of samples&473K&11771\cr
      \specialrule{0em}{2pt}{2pt}
      \hline
      \specialrule{0em}{2pt}{2pt}
      Acquisition periods&10-20(min)&0.6(min)\cr
      \specialrule{0em}{2pt}{2pt}
      \bottomrule
      \end{tabular}
  \end{threeparttable}
  \end{table*}

\begin{table*}
\centering
\caption{Classes and proportions of the OPPORTUNITY dataset}
\setlength{\tabcolsep}{3pt}
\begin{tabular}{cccc}
\hline
\specialrule{0em}{2pt}{2pt}
Class&
Proportion&
Class&
Proportion\\
\specialrule{0em}{2pt}{2pt}
\hline
\specialrule{0em}{1pt}{1pt}
Open Door 1/2&
1.87\%/1.26\%&
Open Fridge&
1.60\%\\
Close Door 1/2&
6.15\%/1.54\%&
Close Fridge&
0.79\%\\
Open Dishwasher&
1.85\%&
Close Dishwasher&
1.32\%\\
Open Drawer 1/2/3&
1.09\%/1.64\%/0.94\%&
Clean Table&
1.23\%\\
Close Drawer 1/2/3&
0.87\%1.69\%/2.04\%&
Drink from Cup&
1.07\%\\
Toggle Switch&
0.78\%&
NULL&
72.28\%\\
\specialrule{0em}{1pt}{1pt}
\hline
\end{tabular}
\label{tab:OPPORTUNITY_DATASET}
\end{table*}

\begin{table*}
\centering
\caption{Classes and proportions of the UniMiB-SHAR dataset}
\setlength{\tabcolsep}{3pt}
\begin{tabular}{cccc}
\hline
\specialrule{0em}{2pt}{2pt}
Class&
Proportion&
Class&
Proportion\\
\specialrule{0em}{2pt}{2pt}
\hline
\specialrule{0em}{1pt}{1pt}
StandingUpfromSitting&
1.30\%&
Walking&
14.77\%\\
StandingUpfromLaying&
1.83\%&
Running&
16.86\%\\
LyingDownfromStanding&
2.51\%&
Going Up&
7.82\%\\
Jumping&
6.34\%&
Going Down&
11.25\%\\
F(alling) Forward&
4.49\%&
F and Hitting Obstacle&
5.62\%\\
F Backward&
4.47\%&
Syncope&
4.36\%\\
F Right&
4.34\%&
F with ProStrategies&
4.11\%\\
F Backward SittingChair&
3.69\%&
F Left&
4.54\%\\
Sitting Down&
1.70\%&
&
\\
\specialrule{0em}{1pt}{1pt}
\hline
\end{tabular}
\label{tab:UniMiB-SHAR_DATASET}
\end{table*}

The OPPORTUNITY dataset and UniMiB-SHAR dataset are collected from real environment. The two datasets have their own characteristic and contain different sensors, the UniMiB-SHAR dataset only contains the accelerometer data, it has low power cost. The OPPORTUNITY dataset combines accelerometers, gyroscopes and magnetic sensors data, and it can provide accurate limb orientation.

\subsection{Baseline}
We compared our proposed ARN method against some classic or state-of-the-art activity recognition methods. We roughly divided these methods into categories: conventional recognition methods include HC~\cite{b22}, CBH~\cite{b23}, CBS~\cite{b24}. The learning-based methods include AE~\cite{b25}, MLP~\cite{b26}, CNN~\cite{b15}, LSTM~\cite{b27}, Hybrid~\cite{b28}, ResNet~\cite{b21}. As in conventional methods, we use hand-crafted features, readers can find more details in~\cite{b38}. For learning-based methods, we use raw activity data as input. Follow by~\cite{b38}, the hyper-parameters of these learning-based baseline models except ResNet\footnote{The hyper-parameters used by ResNet are the hyper-parameters used in one of the path in the proposed ARN model.} for the OPPORTUNITY and UniMiB-SHAR datasets are provided in Table~\ref{tab:learning-basedHyperParamO}.

\begin{table}[tp]
\centering
\begin{threeparttable}
  \centering
  \caption{Hyper-parameters of the learning-based methods on the OPPORTUNITY dataset and UniMiB-SHAR dataset.}
  \label{tab:learning-basedHyperParamO}
    \begin{tabular}{cc}
    \toprule
    Model&parameters\cr
    \midrule
    \specialrule{0em}{2pt}{2pt}
    \multirow{2}*{AE}&5000$^a$\\
                     &5000$^a$\\
    \specialrule{0em}{2pt}{2pt}
    \multirow{3}*{MLP}&2000$^a$\\
                      &2000$^a$\\
                      &2000$^a$\\
    \multirow{4}*{CNN}&((11,1),(1,1),50,(2,1))$^b$\\
                      &((10,1),(1,1),40,(3,1))$^b$\\
                      &((6,1),(1,1),30,(1,1))$^b$\\
                      & 1000$^a$\\
    \specialrule{0em}{2pt}{2pt}
    \multirow{3}*{LSTM}&(64,600)$^c$\\
                       &(64,600)$^c$\\
                       & 512$^a$\\
    \specialrule{0em}{2pt}{2pt}
    \multirow{4}*{Hybrid}&((11,1),(1,1),50,(2,1))$^b$\\
                         &(27,600)$^c$\\
                         &(27,600)$^c$\\
                         & 512$^a$\\
    \specialrule{0em}{2pt}{2pt}
    \bottomrule
    \end{tabular}
    \begin{tablenotes}
    \footnotesize
        \item[$^a$]	dense units.\\
        \item[$^b$] ((kernel size), (siding stride), num of kernels, (Pool)).\\
        \item[$^c$] (number of cells, output-dim).
    \end{tablenotes}
\end{threeparttable}
\end{table}

\subsection{Implementation and Setting}
Our ARN model is implemented in TensorFlow~\cite{b45}, a system that transfers complex data structures to artificial intelligence neural networks for analysis and processing. The computing platform is equipped with an Intel 2$\times$ Intel E5-2600 CPU, 128G RAM, and a NVIDIA TITAN Xp 12G GPU. The model is trained using the ADADELTA gradient decent algorithm with default parameters (i.e., initial learning rate of 1), for 50 epoches. The batch size is set to 128. The hyper-parameters of the proposed model are provides in Table~\ref{tab:model_para}.

\noindent\textbf{Sliding Time Window Size:} The length of the sliding window $T$ is an important hyper-parameter of the proposed model. As in baseline methods, we carried out two more comparative studies using $T=32\ \ (approximately\ \ 1s)$, $T=64\ \ (approximately\ \ 2s)$ and $T=96\ \ (approximately\ \ 3s)$. For the proposed model, we use $T=32$ or $T=64$ as the hyper-parameter of the narrow path and $T=64$ or $T=96$ as the hyper-parameter of the wide path, respectively.

\subsection{Performance Measure}
ADL datasets are often highly unbalanced. The OPPORTUNITY dataset is extremely imbalanced, as the NULL class represents more than 75\% of the recorded data. For this dataset, the overall classification accuracy is not an appropriate measure of performance, because the activity recognition rate of the majority classes might skew the performance statistics to the detriment of the least represented classes. As a result, many previous researches such as~\cite{b30} show the use of an evaluation metric independent of the class repartition---$F1$-score. The $F1$-score combines two measures: the \textbf{precision} $p$ and the \textbf{recall} $r$: $p$ is the number of correct positive examples divided by the number of all positive examples returned by the classifier, and $r$ is the number of correct positive results divided by the number of all positive samples. The $F1$-score is the harmonic average of $p$ and $r$, where the best value is at 1 and worst at 0. In this paper, we use an additional evaluation metric to make the comparison with them easier: the weighted $F1$-Score (Sum of class $F1$-scores, weighted by the class proportion):
 \begin{equation}\label{eq:weight_F1-score}
F_w=2\sum_G{w_g\frac{p_g\cdot r_g}{p_g+r_g}},
 \end{equation}
where $w_g=N_g/N_{total}$ and $N_g$ is the number of samples in class $g$, and $N_{total}$ is the total number of samples.

\begin{table*}[tp]
\centering
\begin{threeparttable}
  \centering
  \caption{Weighted $F_1$-score performances of different methods on the \textbf{OPPORTUNITY} and \textbf{UniMiB-SHAR} datasets. (n) and (w) denote the narrow and wide path, respectively.}
  \label{tab:f1-3sensors}
    \begin{tabular}{lccc}
    \toprule
    \specialrule{0em}{3pt}{3pt}
    Method&$T$ (time window)&OPPORTUNITY&UniMiB-SHAR\cr
    \specialrule{0em}{3pt}{3pt}
    \midrule
    \specialrule{0em}{2pt}{2pt}
    \multirow{3}*{HC~\cite{b22}}&32&84.95&22.83\\
                         &64&85.56&22.19\\
                         &96&85.69&21.96\\
    \specialrule{0em}{2pt}{2pt}
    \multirow{3}*{CBH~\cite{b23}}&32&84.37&64.51\\
                         &64&85.21&65.03\\
                         &96&84.66&64.36\\
    \specialrule{0em}{2pt}{2pt}
    \multirow{3}*{CBS~\cite{b24}}&32&85.53&67.54\\
                         &64&86.01&67.97\\
                         &96&85.39&67.36\\
    \specialrule{0em}{2pt}{2pt}
    \hline
    \specialrule{0em}{2pt}{2pt}
    \multirow{3}*{AE~\cite{b25}}&32&82.87&68.37\\
                         &64&84.54&68.24\\
                         &96&83.39&68.39\\
    \specialrule{0em}{2pt}{2pt}
    \multirow{3}*{MLP~\cite{b26}}&32&87.32&73.33\\
                         &64&87.34&75.36\\
                         &96&86.65&74.82\\
    \specialrule{0em}{2pt}{2pt}
    \multirow{3}*{CNN~\cite{b15}}&32&87.51&74.01\\
                         &64&88.03&73.04\\
                         &96&87.62&73.36\\
    \specialrule{0em}{2pt}{2pt}
    \multirow{3}*{LSTM~\cite{b27}}&32&85.33&69.24\\
                         &64&86.89&69.49\\
                         &96&86.21&68.81\\
    \specialrule{0em}{2pt}{2pt}
    \multirow{3}*{Hybrid~\cite{b28}}&32&87.91&73.19\\
                         &64&88.17&73.22\\
                         &96&87.67&72.26\\
    \specialrule{0em}{2pt}{2pt}
    \hline
    \specialrule{0em}{2pt}{2pt}
    \multirow{3}*{ResNet~\cite{b21}}&32&88.91&76.19\\
                         &64&89.17&76.22\\
                         &96&87.67&75.26\\
    \specialrule{0em}{2pt}{2pt}
    \hline
    \specialrule{0em}{2pt}{2pt}
    ARN&32-96 (n)-(w)&\textbf{90.29}&\textbf{76.39}\cr
    \bottomrule
    \end{tabular}
\end{threeparttable}
\end{table*}

\subsection{Results and discussions}
In this section, we present and discuss the results. To get insight into how these methods are applied to the domain, we show the performance of these methods and evaluate some key parameters.

\begin{table*}[tp]
\centering
\begin{threeparttable}
  \centering
  \caption{Weighted $F_1$-score performances comparison of ARN with the combinations of different lengths of the slide window on the OPPORTUNITY and UniMiB-SHAR datasets. (n) and (w) denote the narrow and wide path, respectively.}
  \label{tab:results}
    \begin{tabular}{cccc}
    \toprule
    Method&$T$ (n)-(w) (time window)&OPPORTUNITY&UniMiB-SHAR\cr
    \midrule
   \specialrule{0em}{2pt}{2pt}
   ARN\_1&32-64&90.21&\textbf{77.23}\cr
    \specialrule{0em}{2pt}{2pt}
    ARN\_2&32-96&\textbf{90.29}&76.39\cr
    \specialrule{0em}{2pt}{2pt}
    ARN\_3&64-96&90.19&76.04\cr
    \bottomrule
    \end{tabular}
\end{threeparttable}
\end{table*}

The weighted $F_1$-score of all models on OPPORTUNITY and UniMiB-SHAR are listed in Table~\ref{tab:f1-3sensors}. Results on these datasets show that the proposed ARN method substantially outperforms all other methods against which it was compared. Compared to conventional recognition methods, such as CBS, the best conventional method achieves an absolute boost of 4.98\%, and 14.65\% corresponding to the OPPORTUNITY dataset and the UniMiB-SHAR dataset, respectively. In addition, most of the learning-based recognition methods outperform the conventional recognition methods. In particular, for OPPORTUNITY dataset, the Hybrid method achieves the best performance among all the learning-based methods. Compared to Hybrid method, our ARN method achieves boosts of 2.4\%. For UniMib-SHAR dataset, the MLP method achieves the best performance among all the learning-based methods. Compared to MLP method, our ARN method achieves boosts of 2.48\%. We also compared a single-path residual network i.e. ResNet, our ARN achieves an absolute boost of 1.26\%, and 1.32\% corresponding to the OPPORTUNITY dataset and the UniMiB-SHAR dataset, respectively.

From the Table~\ref{tab:f1-3sensors}, we can observe that the gap between the learning-based methods and conventional methods is larger on the UniMiB-SHAR dataset than OPPORTUNITY dataset. The reasons are that the sensor channels in OPPORTUNITY dataset are more than those in UniMiB-SHAR dataset. By carefully comparing the performance of the results, we found that our proposed method showed a higher degree of performance improvement when tested on UniMiB-SHAR dataset compared to OPPORTUNITY dataset. This means that our method is effective.

We also observe from the Table~\ref{tab:f1-3sensors} that different lengths of slide time window have an impact on the performance of the activity recognition. The short time window contains too little information. With the growth of the time window, the window contains more and more information, and the accuracy is improved accordingly. But using longer slide time window does not yield better recognition performances~\cite{DBLP:journals/sensors/LiSNKG47}. Most methods perform best in human activity recognition tasks when $T=64$. The reasons are that longer frames potentially contain data related to a higher number of activities, making their majority-labeling more inaccurate.

\subsection{Hyperparameters Evaluation}
\noindent\textbf{Impact of the length of the narrow and wide path selection:}
The model we proposed has two paths, one is narrow path (i.e., the length of slide time window is short), another is wide path (i.e., the length of the slide window is wide). In order to verify the impact of different lengths of slide time windows combinations on the results. We leverage a combination of slide time window of different lengths for comparison experiments (i.e., 32-64, 32-96, 64-96). ARN\_1, ARN\_2 and ARN\_3 mean the combination of slide time window are 32-64, 32-96 and 64-96, respectively. We carried out experiments on the two datasets. The weighted $F_1$-score results are shown in Table~\ref{tab:results}. From Table~\ref{tab:results}, we can observe that on the OPPORTUNITY dataset, the ARN\_2 outperforms the ARN\_1 and ARN\_3, the minimum image size is $63(D)$*$32(T)$ and the accuracy is 90.29\%. For UniMiB-SHAR dataset, the ARN\_1 outperforms the ARN\_2 and ARN\_3, the minimum image size is $3(D)$*$32(T)$ and the accuracy is 76.39\%. The performance gap between the three experiments was very small. This indicates that ARN is a stable model that is not sensitive to the lengths of slide time window.

\section{Conclusions}\label{sec:conc}

In this paper, we propose a novel asymmetric residual network for activity recognition using wearable device data, named ARN. To improve the accuracy of activity recognition, our method consists of two paths. The first path uses a short time window to capture spatial features, and the second path uses a long time window to capture fine temporal features. Unlike other learning-based methods, ARN considers the spatial and temporal features of the data at the same time. It can effectively manage information flow and automatically learn activity feature representation. ARN is an end-to-end network, and the data collected by the wearable device can be directly input into the network. Comprehensive experiments on the two benchmark human activity recognition datasets demonstrate that the ARN outperforms the compared baselines. This method has a good application prospect. However, biometric identification, fingerprint recognition, iris recognition and other technologies have achieved more than 98\% accuracy and are widely used in people’s life. Compared to the practical applications of recognition field in society, the accuracy of  ARN cannot meet. Therefore, there is still a lot of room for progress.

For future work we may research how to extract more fine-gained features by using attention strategy and focus on the research of data dynamic fusion algorithm for maximizing the retention of the data features and obtaining higher recognition accuracy. In addition, we will recognize human dangerous activities, but these activities recognition involves many fine-gained feature extractions. Therefore, we may take advantage of the memory mechanism to design a memory-augmented neural network~\cite{b46} that can learn to find supporting pre-stored clews (i.e., representations of historical activities or others).



\begin{thebibliography}{00}

\bibitem{b1} Sukor, A.S.A.; Zakaria, A.; Rahim, N.A.; Kamarudin, L.M.; Setchi, R.; Nishizaki, H. A hybrid approach of knowledge-driven and data-driven reasoning for activity recognition in smart homes. Journal of Intelligent and Fuzzy Systems 2019, 36, 4177-4188. doi:10.3233/JIFS-169976.

\bibitem{b2} Xiao, Z.; Lim,H.B.; Ponnambalam, L. Participatory Sensing for Smart Cities: A Case Study on Transport Trip Quality Measurement. IEEE Trans. Industrial Informatics 2017, 13, 759-770. doi:10.1109/TII.2017.2678522.

\bibitem{b3} Fortino, G.; Ghasemzadeh, H.; Gravina, R.; Liu, P.X.; Poon, C.C.Y.; Wang, Z. Advances in multi-sensor fusion for body sensor networks: Algorithms, architectures, and applications. Information Fusion 2019, 45, 150-152. doi:10.1016/j.inffus.2018.01.012.

\bibitem{b4} Qiu, J.X.; Yoon, H.; Fearn, P.A.; Tourassi, G.D. Deep Learning for Automated Extraction of Primary Sites From Cancer Pathology Reports. IEEE J. Biomedical and Health Informatics 2018, 22, 244-251.
doi:10.1109/JBHI.2017.2700722.

\bibitem{b5} Oh, I.; Cho, H.; Kim, K. Playing real-time strategy games by imitating human players’ micromanagement skills based on spatial analysis. Expert Syst. Appl. 2017, 71, 192-205. doi:10.1016/j.eswa.2016.11.026.

\bibitem{b6} Lisowska, A.; O’Neil, A.; Poole, I. Cross-cohort Evaluation of Machine Learning Approaches to Fall Detection from Accelerometer Data. Proceedings of the 11th International Joint Conference on Biomedical Engineering Systems and Technologies (BIOSTEC 2018) - Volume 5: HEALTHINF, Funchal, Madeira, Portugal, January 19-21, 2018., 2018, pp. 77-82. doi:10.5220/0006554400770082.

\bibitem{b7} Attal, F.; Mohammed, S.; Dedabrishvili, M.; Chamroukhi, F.; Oukhellou, L.; Amirat, Y. Physical Human Activity Recognition Using Wearable Sensors. Sensors 2015, 15, 31314-31338. doi:10.3390/s151229858.



\bibitem{b9} Bao, L.; Intille, S.S. Activity Recognition from User-Annotated Acceleration Data. Pervasive Computing, Second International Conference, PERVASIVE 2004, Vienna, Austria, April 21-23, 2004, Proceedings, 2004, pp. 1-17. doi:10.1007/978-3-540-24646-6\_1.

\bibitem{b10} Ordóñez, F.J.; Englebienne, G.; de Toledo, P.; van Kasteren, T.; Sanchis, A.; Kröse, B.J.A. In-Home Activity Recognition: Bayesian Inference for Hidden Markov Models. IEEE Pervasive Computing 2014, 13, 67-75. doi:10.1109/MPRV.2014.52.

\bibitem{b11} Yang, Z.; Raymond, O.I.; Sun, W.; Long, J. Deep Attention-Guided Hashing. IEEE Access 2019, 7, 11209-11221. doi:10.1109/ACCESS.2019.2891894.

\bibitem{b12} Sarkar, A.; Dasgupta, S.; Naskar, S.K.; Bandyopadhyay, S. Says Who? Deep Learning Models for Joint Speech Recognition, Segmentation and Diarization. 2018 IEEE International Conference on Acoustics, Speech and Signal Processing, ICASSP 2018, Calgary, AB, Canada, April 15-20, 2018, 2018, pp. 5229-5233. doi:10.1109/ICASSP.2018.8462375.

\bibitem{b13} Kumar, A.; Irsoy, O.; 413 Ondruska, P.; Iyyer, M.; Bradbury, J.; Gulrajani, I.; Zhong, V.; Paulus, R.; Socher, R. Ask Me Anything: Dynamic Memory Networks for Natural Language Processing. Proceedings of the 33nd International Conference on Machine Learning, ICML 2016, New York City, NY, USA, June 19-24, 2016, 2016, pp. 1378-1387.

\bibitem{b14} Shi, Z.; Zhang, J.A.; Xu, R.; Fang, G. Human Activity Recognition Using Deep Learning Networks with Enhanced Channel State Information. IEEE Globecom Workshops, GC Wkshps 2018, Abu Dhabi, United Arab Emirates, December 9-13, 2018, 2018, pp. 1-6. doi:10.1109/GLOCOMW.2018.8644435.

\bibitem{b15} Xu, W.; Pang, Y.; Yang, Y.; Liu, Y. Human Activity Recognition Based On Convolutional Neural Network. 24th International Conference on Pattern Recognition, ICPR 2018, Beijing, China, August 20-24, 2018, 2018, pp. 165-170. doi:10.1109/ICPR.2018.8545435.

\bibitem{b16} Hammerla, N.Y.; Halloran, S.; Plötz, T. Deep, Convolutional, and Recurrent Models for Human Activity Recognition Using Wearables. Proceedings of the Twenty-Fifth International Joint Conference on Artificial Intelligence, IJCAI 2016, New York, NY, USA, 9-15 July 2016, 2016, pp. 1533-1540.

\bibitem{b17} Neverova, N.;Wolf, C.; Lacey, G.; Fridman, L.; Chandra, D.; Barbello, B.; Taylor, G.W. Learning Human Identity From Motion Patterns. IEEE Access 2016, 4, 1810-1820. doi:10.1109/ACCESS.2016.2557846.

\bibitem{b18} Yang, J.; Nguyen, M.N.; San, P.P.; Li, X.; Krishnaswamy, S. Deep Convolutional Neural Networks on Multichannel Time Series for Human Activity Recognition. Proceedings of the Twenty-Fourth International Joint Conference on Artificial Intelligence, IJCAI 2015, Buenos Aires, Argentina, July 25-31, 2015, 2015, pp. 3995-4001.

\bibitem{b19} Yang, Z.; Raymond, O.I.; Zhang, C.; Wan, Y.; Long, J. DFTerNet: Towards 2-bit Dynamic Fusion Networks for Accurate Human Activity Recognition. IEEE Access 2018, 6, 56750-56764. doi:10.1109/ACCESS.2018.2873315.

\bibitem{b20} Ronao, C.A.; Cho, S. Deep Convolutional Neural Networks for Human Activity Recognition with Smartphone Sensors. Neural Information Processing - 22nd International Conference, ICONIP 2015, Istanbul, Turkey, November 9-12, 2015, Proceedings, Part IV, 2015, pp. 46-53. doi:10.1007/978-3-319-26561-2\_6.

\bibitem{b21} He, K.; Zhang, X.; Ren, S.; Sun, J. Deep Residual Learning for Image Recognition. 2016 IEEE Conference on Computer Vision and Pattern Recognition, CVPR 2016, Las Vegas, NV, USA, June 27-30, 2016, 2016, pp. 770-778. doi:10.1109/CVPR.2016.90.

\bibitem{b22} Ramamurthy, S.R.; Roy, N. Recent trends in machine learning for human activity recognition - A survey. Wiley Interdiscip. Rev. Data Min. Knowl. Discov. 2018, 8. doi:10.1002/widm.1254.

\bibitem{b23} Shirahama, K.; Grzegorzek,M. On the Generality of Codebook Approach for Sensor-Based Human Activity Recognition 2017.

\bibitem{b24} van Gemert, J.C.; Veenman, C.J.; Smeulders, A.W.M.; Geusebroek, J. VisualWord Ambiguity. IEEE Trans. Pattern Anal. Mach. Intell. 2010, 32, 1271-1283. doi:10.1109/TPAMI.2009.132.

\bibitem{b25} Gu, F.; Khoshelham, K.; Valaee, S.; Shang, J.; Zhang, R. Locomotion Activity Recognition Using Stacked Denoising Autoencoders. IEEE Internet of Things Journal 2018, 5, 2085-2093. doi:10.1109/JIOT.2018.2823084.

\bibitem{b26} Ioffe, S.; Szegedy, C. Batch Normalization: Accelerating Deep Network Training by Reducing Internal Covariate Shift. Proceedings of the 32nd International Conference on Machine Learning, ICML 2015, Lille, France, 6-11 July 2015, 2015, pp. 448-456.

\bibitem{b27} Milenkoski, M.; Trivodaliev, K.; Kalajdziski, S.; Jovanov, M.; Stojkoska, B.R. Real time human activity recognition on smartphones using LSTM networks. 41st International Convention on Information and Communication Technology, Electronics and Microelectronics, MIPRO 2018, Opatija, Croatia, May 21-25, 2018, 2018, pp. 1126-1131. doi:10.23919/MIPRO.2018.8400205.

\bibitem{b28} Meng, B.; Liu, X.; Wang, X. Human action recognition based on quaternion spatial-temporal convolutional neural network and LSTM in RGB videos. Multimedia Tools Appl. 2018, 77, 26901-26918. doi:10.1007/s11042-018-5893-9.

\bibitem{b29} Han, J.; Kamber, M.; Pei, J. Data Mining: Concepts and Techniques, 3rd edition; Morgan Kaufmann, 2011.

\bibitem{b30} Morales, F.J.O.; Roggen, D. Deep Convolutional and LSTM Recurrent Neural Networks for Multimodal Wearable Activity Recognition. Sensors 2016, 16, 115. doi:10.3390/s16010115.

\bibitem{b31} Baldi, P.; Sadowski, P.J. A theory of local learning, the learning channel, and the optimality of backpropagation. Neural Networks 2016, 83, 51-74. doi:10.1016/j.neunet.2016.07.006.

\bibitem{b32} Nair, V.; Hinton, G.E. Rectified Linear Units Improve Restricted Boltzmann Machines. Proceedings of the 27th International Conference on Machine Learning (ICML-10), June 21-24, 2010, Haifa, Israel, 2010, pp. 807-814.

\bibitem{b33} Glorot, X.; Bordes, A.; Bengio, Y. Deep Sparse Rectifier Neural Networks. Proceedings of the Fourteenth International Conference on Artificial Intelligence and Statistics, AISTATS 2011, Fort Lauderdale, USA, April 11-13, 2011, 2011, pp. 315-323.

\bibitem{b34} Pascanu, R.; Mikolov, T.; Bengio, Y. On the difficulty of training recurrent neural networks. Proceedings of the 30th International Conference on Machine Learning, ICML 2013, Atlanta, GA, USA, 16-21 June 2013, 2013, pp. 1310-1318.

\bibitem{b35} Carreira, J.; Zisserman, A. Quo Vadis, Action Recognition? A New Model and the Kinetics Dataset. 2017 IEEE Conference on Computer Vision and Pattern Recognition, CVPR 2017, Honolulu, HI, USA, July 21-26, 2017, 2017, pp. 4724-4733. doi:10.1109/CVPR.2017.502.

\bibitem{b36} Feichtenhofer, C.; Pinz, A.; Wildes, R.P. Spatiotemporal Residual Networks for Video Action Recognition. Advances in Neural Information Processing Systems 29: Annual Conference on Neural Information Processing Systems 2016, December 5-10, 2016, Barcelona, Spain, 2016, pp. 3468-3476.

\bibitem{b37} Wang, X.; Girshick, R.B.; Gupta, A.; He, K. Non-Local Neural Networks. 2018 IEEE Conference on Computer Vision and Pattern Recognition, CVPR 2018, Salt Lake City, UT, USA, June 18-22, 2018, 2018, pp. 7794-7803. doi:10.1109/CVPR.2018.00813.

\bibitem{b38} Li, F.; Shirahama, K.; Nisar, M.A.; Köping, L.; Grzegorzek, M. Comparison of Feature Learning Methods for Human Activity Recognition Using Wearable Sensors. Sensors 2018, 18, 679. doi:10.3390/s18020679.

\bibitem{b39} Feichtenhofer, C.; Fan, H.; Malik, J.; He, K. SlowFast Networks for Video Recognition. CoRR 2018, abs/1812.03982, [1812.03982].

\bibitem{b40} Roggen, D.; Calatroni, A.; Rossi, M.; Holleczek, T.; Förster, K.; Tröster, G.; Lukowicz, P.; Bannach, D.; Pirkl, G.; Ferscha, A.; Doppler, J.; Holzmann, C.; Kurz, M.; Holl, G.; Chavarriaga, R.; Sagha, H.; Bayati, H.; Creatura, M.; del R. Millán, J. Collecting complex activity datasets in highly rich networked sensor environments. Seventh International Conference on Networked Sensing Systems, INSS 2010, Kassel, Germany, June 15-18, 2010, 2010, pp. 233-240. doi:10.1109/INSS.2010.5573462.

\bibitem{b41} Kwapisz, J.R.; Weiss, G.M.; Moore, S. Activity recognition using cell phone accelerometers. SIGKDD Explorations 2010, 12, 74-82. doi:10.1145/1964897.1964918.

\bibitem{b42} Micucci, D.; Mobilio, M.; Napoletano, P. UniMiB SHAR: a new dataset for human activity recognition using acceleration data from smartphones. CoRR 2016, abs/1611.07688, [1611.07688].

\bibitem{b43} Ba{\~{n}}os, O.; Garc{\'{\i}}a, R.; Terriza, J.A.H.; Damas, M.; Pomares, H.; Ruiz, I.R.; Saez, A.; Villalonga, C. mHealthDroid: A Novel Framework for Agile Development of Mobile Health Applications. Ambient Assisted Living and Daily Activities - 6th International Work-Conference, IWAAL 2014, Belfast, UK, December 2-5, 2014. Proceedings, 2014, pp. 91-98. doi:10.1007/978-3-319-13105-4\_14.

\bibitem{b44} Reiss, A.; Stricker, D. Creating and benchmarking a new dataset for physical activity monitoring. The 5th International Conference on PErvasive Technologies Related to Assistive Environments, PETRA 2012, Heraklion, Crete, Greece, June 6-9, 2012, 2012, p. 40. doi:10.1145/2413097.2413148.

\bibitem{b45} Abadi, M.; Agarwal, A.; Barham, P.; Brevdo, E.; Chen, Z.; Citro, C.; Corrado, G.S.; Davis, A.; Dean, J.; Devin, M.; Ghemawat, S.; Goodfellow, I.J.; Harp, A.; Irving, G.; Isard, M.; Jia, Y.; Józefowicz, R.; Kaiser, L.; Kudlur, M.; Levenberg, J.; Mané, D.; Monga, R.; Moore, S.; Murray, D.G.; Olah, C.; Schuster, M.; Shlens, J.; Steiner, B.; Sutskever, I.; Talwar, K.; Tucker, P.A.; Vanhoucke, V.; Vasudevan, V.; Viégas, F.B.; Vinyals, O.; Warden, P.;Wattenberg, M.;Wicke, M.; Yu, Y.; Zheng, X. TensorFlow: Large-Scale Machine Learning on Heterogeneous Distributed Systems. CoRR 2016, abs/1603.04467, [1603.04467].

\bibitem{b46} Yang, Z.; He, X.; Gao, J.; Deng, L.; Smola, A.J. Stacked Attention Networks for Image Question Answering. In Proceedings of the 2016 IEEE Conference on Computer Vision and Pattern Recognition, Las Vegas, NV, USA, 27-30 June 2016; pp. 21-29, doi:10.1109/CVPR.2016.10.

\end{thebibliography}
\end{document}